\documentclass[conference]{IEEEtran}
\usepackage{cite}

\ifCLASSINFOpdf
  \usepackage[pdftex]{graphicx}
  \usepackage{tikz}
  \usepackage{float}
  \usepackage{bm}
  \usepackage{placeins}
\else
\fi
\usepackage{amsmath}
\begin{document}
%
\title{Addressing Relative Pose Impact on UWB Localization: Dataset Introduction and Analysis}

\author{\IEEEauthorblockN{Jun Hyeok Choe$^{1}$}
\IEEEauthorblockA{Aerospace Engineering\\
Inha University\\
Incheon, 21999, Korea\\
Email: jhchoe@inha.ac.kr}
\and
\IEEEauthorblockN{Inwook Shim$^{\dag}$}
\IEEEauthorblockA{Smart Mobility Engineering\\
Inha University\\
Incheon, 21999, Korea\\
Email: iwshim@inha.ac.kr}
}

\maketitle

\begin{abstract}
UWB has recently gained new attention as an auxiliary sensor in the field of robot localization due to its compactness and ease of distance measurement. Consequently, various UWB-related localization and dataset research have increased. Despite this broad interest, there is a lack of UWB datasets that thoroughly analyze the performance of UWB ranging measurement. To address this issue, our paper introduces a UWB dataset that examines UWB relative pose factors affecting ranging measurement. To the best of our knowledge, our dataset is the first to analyze these factors while rigorously providing precise ground-truth UWB poses. The dataset is accessible at \tt{https://github.com/cjhhalla/RCV\_uwb\_dataset}.
\end{abstract}


%
\IEEEpeerreviewmaketitle

\section{Introduction}
Localization is a fundamental module for various applications, such as robotics and autonomous driving. With centimeter-level accuracy and affordable cost, Ultra-Wideband~(UWB) has gained significant interest for localization application~\cite{Comprehensive-Overview-on-UWB-radar}. Moreover, UWB provides robust measurements in adverse weather conditions such as rain, snow, and fog compared to vision and LiDAR~\cite{bad_weather_UWB}. Due to its robustness to environmental conditions, UWB is emerging as a new sensor for localization in robotics systems. The UWB system achieves localization by attaching tags to mobile robots, which communicate with pre-installed anchors in the environment.

To promote research on UWB-based localization, various datasets have been released recently~\cite{UTIL,NTU,queralta2020uwb,MURP}. These datasets have enabled numerous studies~\cite{LearningBase,fishberg2022multiagent,Finding_the_right} addressing key factors affecting UWB localization performance, such as Non-Line-of-Sight~(NLOS) and multi-path interference. However, despite these advancements, there is still a lack of publicly available datasets suitable for evaluating the impact of relative pose between tags and anchors, which is one of the major factors contributing to performance degradation. 

In this paper, to tackle the UWB relative pose issue, we introduce a UWB localization dataset with ground-truth relative poses and provide analysis results on pose-dependent errors and biases according to the changing relative pose between tags and anchors. To achieve this, we design new rigid body frames for the drone and UWB tags and anchors. Using the motion capture system, conducting experiments across various scenarios.

\vspace{-3mm}
\section{Related work}
\begin{figure}[t]
  \centering
  \includegraphics[width=0.5\textwidth]{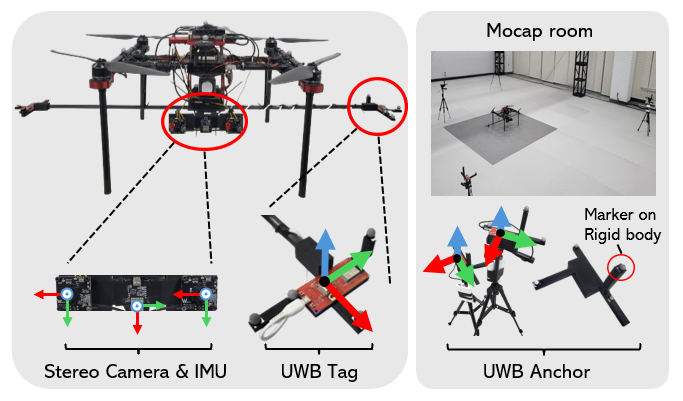}
  \caption{Two synchronized monocular global shutter cameras, along with an IMU and a DW1000-based UWB tag extended on a rod, are mounted on the drone. The UWB is equipped with a rigid body frame, which determines the UWB's coordinate system. In the coordinate system, the UWB's origin is the center of the UWB device chip. Additionally, the coordinate axes are constructed following the right-hand rule. We adopt a single tag to avoid obstructing the signal path between the ranging measurement pairs. The upper right figure shows our experimental environment.}
  \label{overview}
\end{figure}
\begin{figure*}[t]
  \centering
  \includegraphics[width=1\textwidth]{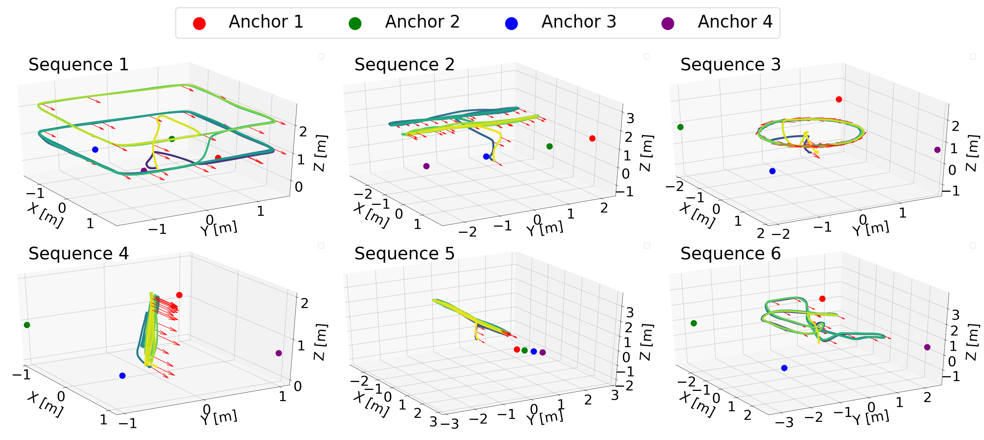}
  \caption{ 
    Each trajectory sequence is recorded from the drone's takeoff to its landing with different anchor placements. Darker and brighter colors indicate takeoff and landing, respectively.
    The static anchors have their respective heading directions and are at the same height in Sequences 1 and 2, whereas in the other sequences, the anchor heights vary. The red arrows in the trajectory demonstrate the drone's heading direction. 
    }
  \label{Sequence}
\end{figure*}
\begin{figure}[t]
  \includegraphics[width=\linewidth]{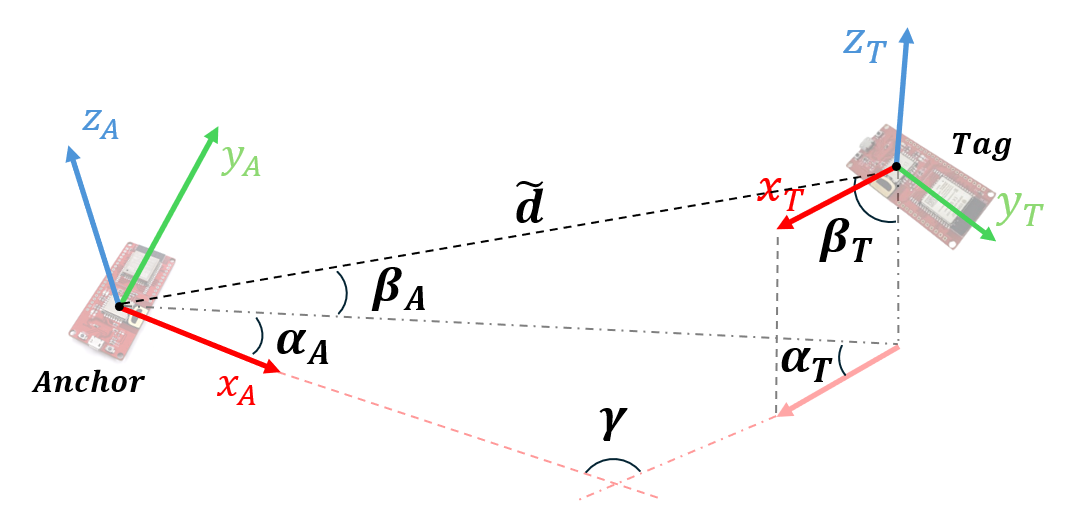}
  \caption{Notation of the relative pose between tag~$T$ and anchor~$A$: ranging measurement $\tilde{d}$, azimuth angle $\alpha_A$ and $\alpha_B$, elevation angle $\beta_A$ and $\beta_B$, and the difference angle $\gamma$, which represents the difference in heading angles.}
  \label{anchor_setup}
\end{figure}
Several notable public localization datasets, including UWB data, address diverse scenarios incorporating ground-truth robot pose information.
The UTIL~\cite{UTIL} focuses on both Non-Line-of-Sight~(NLOS) and Line-of-Sight~(LOS) conditions with dynamic obstacles in cluttered environments, but it overlooks the UWB relative pose issue.
The NTU VIRAL~\cite{NTU} captures UWB ranging measurements in outdoor environments using other vision sensors. Although they provide valuable data, this dataset lacks information on the orientation of the anchors.
The MURP~\cite{MURP} is the most similar to our work, as it also focuses on the relative pose between UWB tags and anchors. However, Their experiments involve placing too many UWB devices in a limited area, inevitably leading to multi-path interference among the UWB devices~\cite{fishberg2022multiagent}. These make it difficult to analyze the UWB relative pose issue independently.

To the best of our knowledge, no dataset rigorously addresses ranging measurement issues based on UWB's relative pose. Therefore, our dataset will be crucial role in addressing the challenges associated with UWB relative pose issue.

\section{Dataset characteristics}
\subsection{System setup}
An overview of our system setup is illustrated in Fig.~\ref{overview}. The tag and anchor frames are custom-designed, with markers attached to accurately capture the 6-DOF pose using the motion capture system. Note that only a single tag is utilized to collect data in the dataset to avoid interference in ranging measurements between tags. Antenna delay calibration between UWB tag and anchors is performed using the method described in \cite{decawave_aps014}.
Our dataset also includes stereo vision and inertial sensor data for future research. The stereo and inertial sensors are mounted on the bottom plate of the drone, facing forward. Each camera is time-synchronized using an external trigger. Stereo camera intrinsics and camera-IMU extrinsic parameters are obtained using Kalibr~\cite{kalibr}.

\subsection{Dataset sequence}

Our dataset is collected at the motion capture room~(see Fig.~\ref{overview}), which gives precise $6$-D pose information. Each sequence, illustrated in  Fig.~\ref{Sequence}, is designed to examine the impact of ranging measurement according to the relative pose of UWBs, including azimuth angle ($\alpha_A$, $\alpha_T$), elevation angle ($\beta_A$, $\beta_T$), difference angle ($\gamma$), and UWB ranging measurement ($\tilde{d}$) as shown in Fig.~\ref{anchor_setup}.

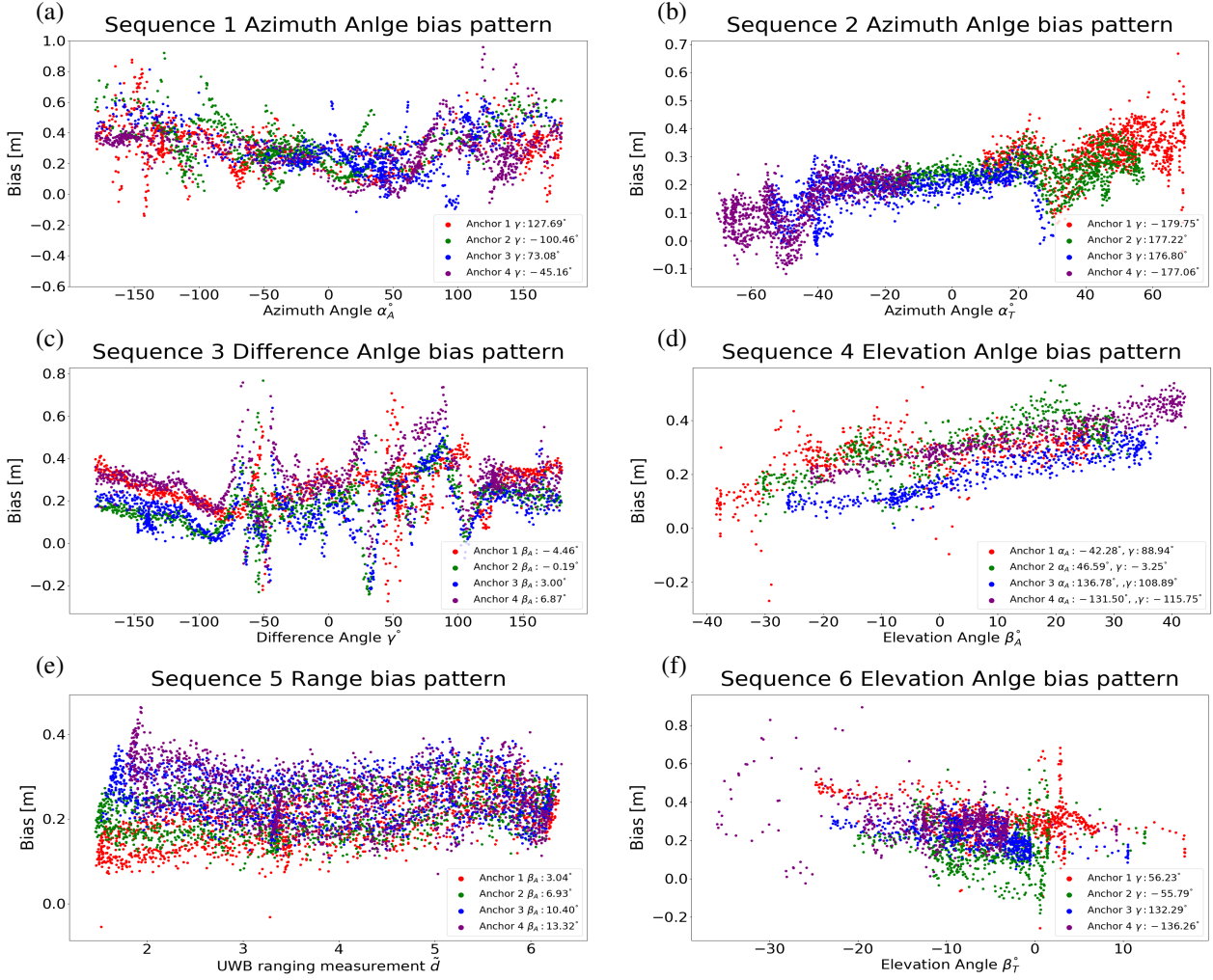
\begin{figure*}[t]
  \centering
  \hspace{-2.8em}
  \begin{tabular}{cc}
    \begin{tikzpicture}
      \node[inner sep=0] at (0,0) {\includegraphics[width=0.45\textwidth, height=0.18\textheight]{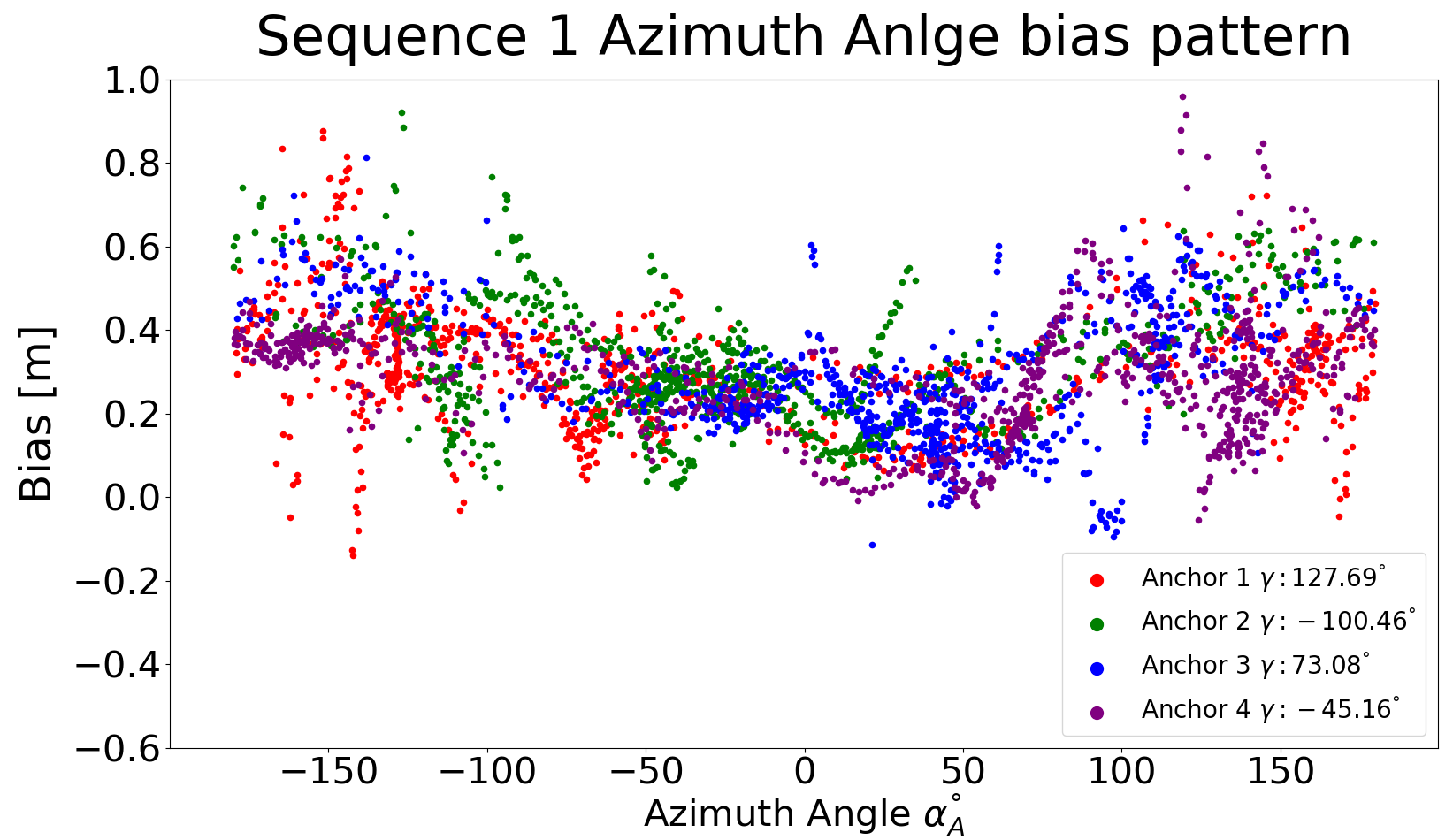}};
      \node[above left, inner sep=0em, xshift=2.5em, yshift=-0.5em] at (current bounding box.north west) {(a)};
    \end{tikzpicture} &
    \begin{tikzpicture}
      \node[inner sep=0] at (0,0) {\includegraphics[width=0.45\textwidth, height=0.18\textheight]{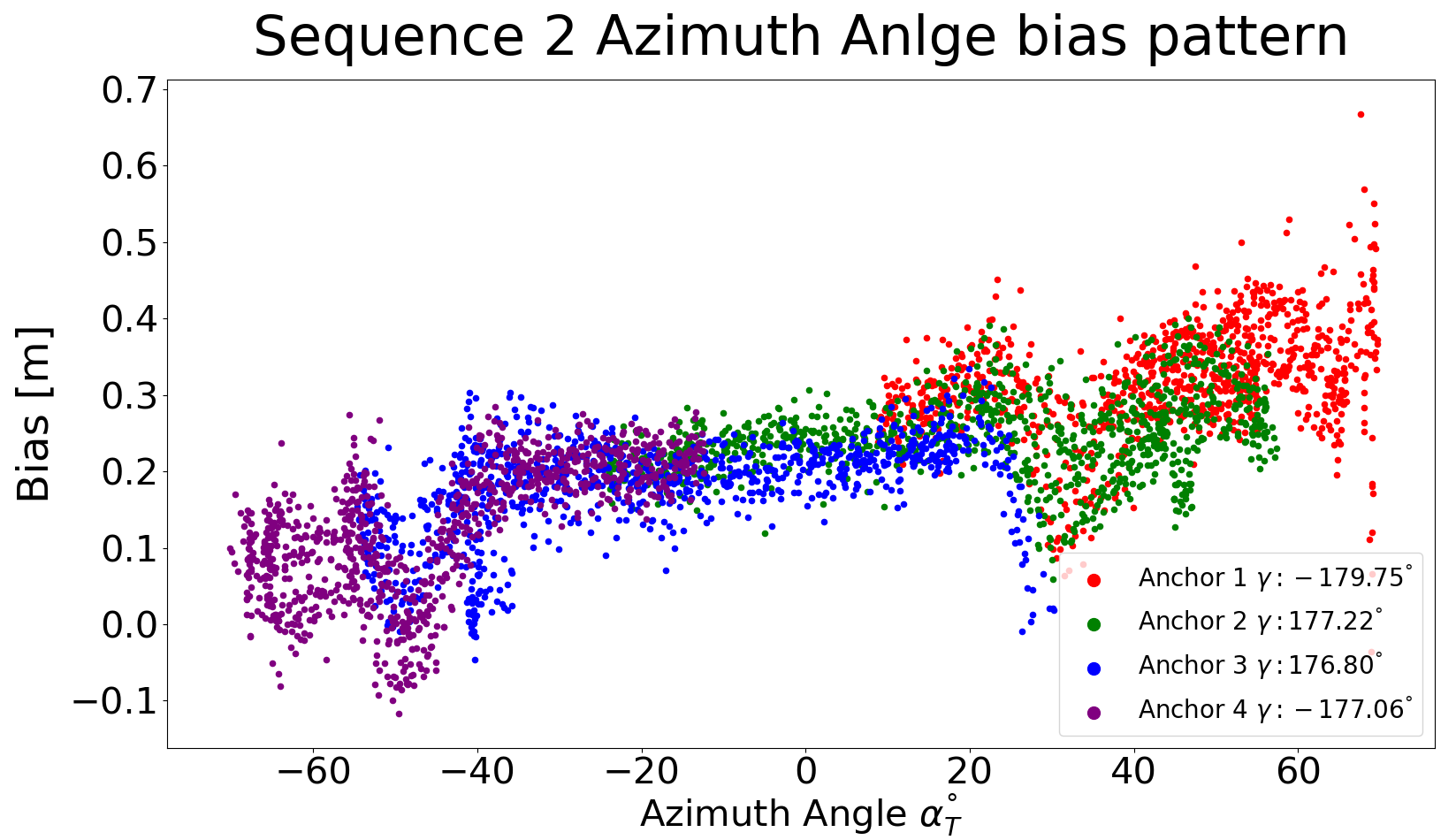}};
      \node[above left, inner sep=0em, xshift=2.5em, yshift=-0.5em] at (current bounding box.north west) {(b)};
    \end{tikzpicture} \\
    \begin{tikzpicture}
      \node[inner sep=0] at (0,0) {\includegraphics[width=0.45\textwidth, height=0.18\textheight]{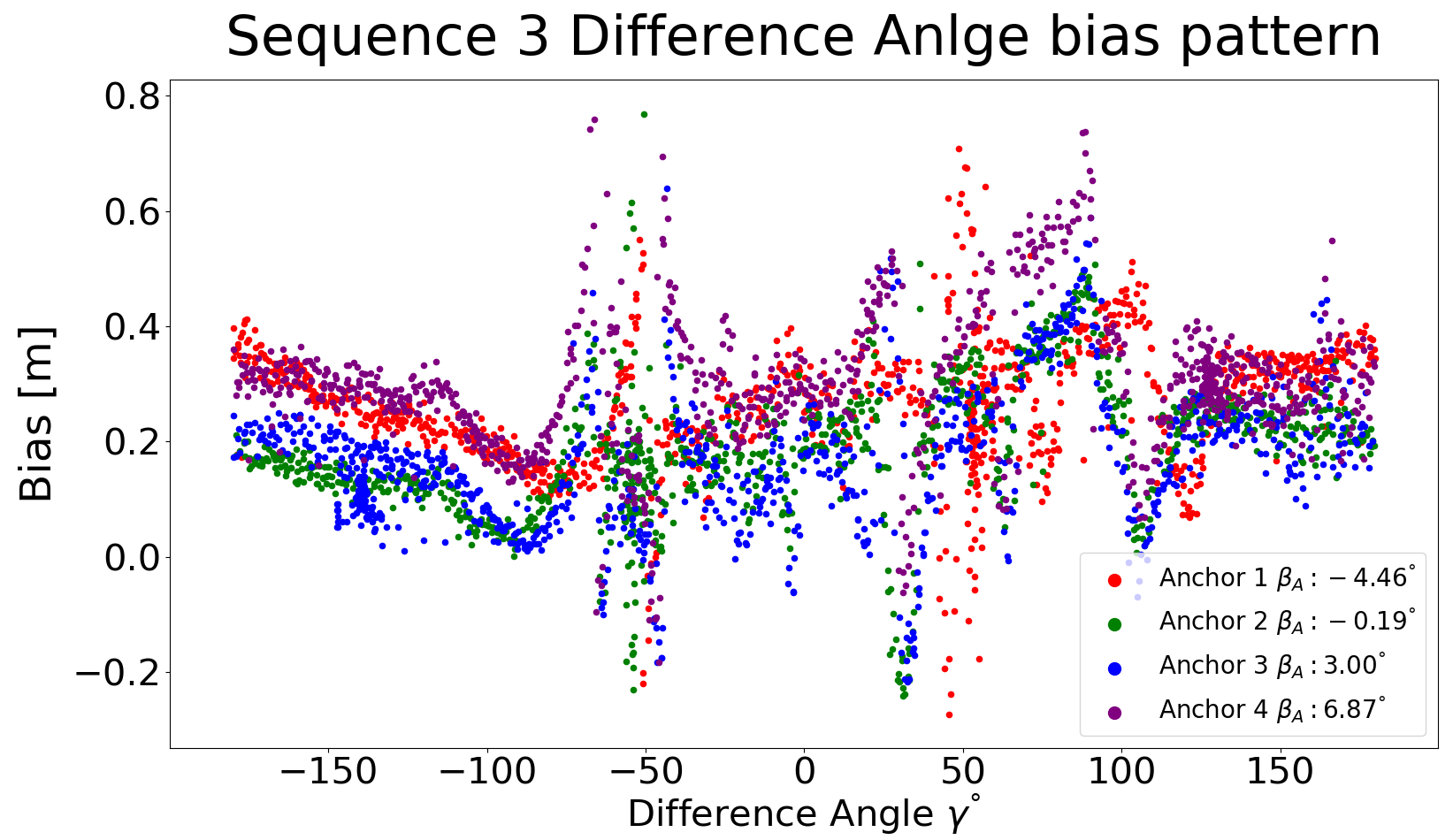}};
      \node[above left, inner sep=0em, xshift=2.5em, yshift=-0.5em] at (current bounding box.north west) {(c)};
    \end{tikzpicture} &
    \begin{tikzpicture}
      \node[inner sep=0] at (0,0) {\includegraphics[width=0.45\textwidth, height=0.18\textheight]{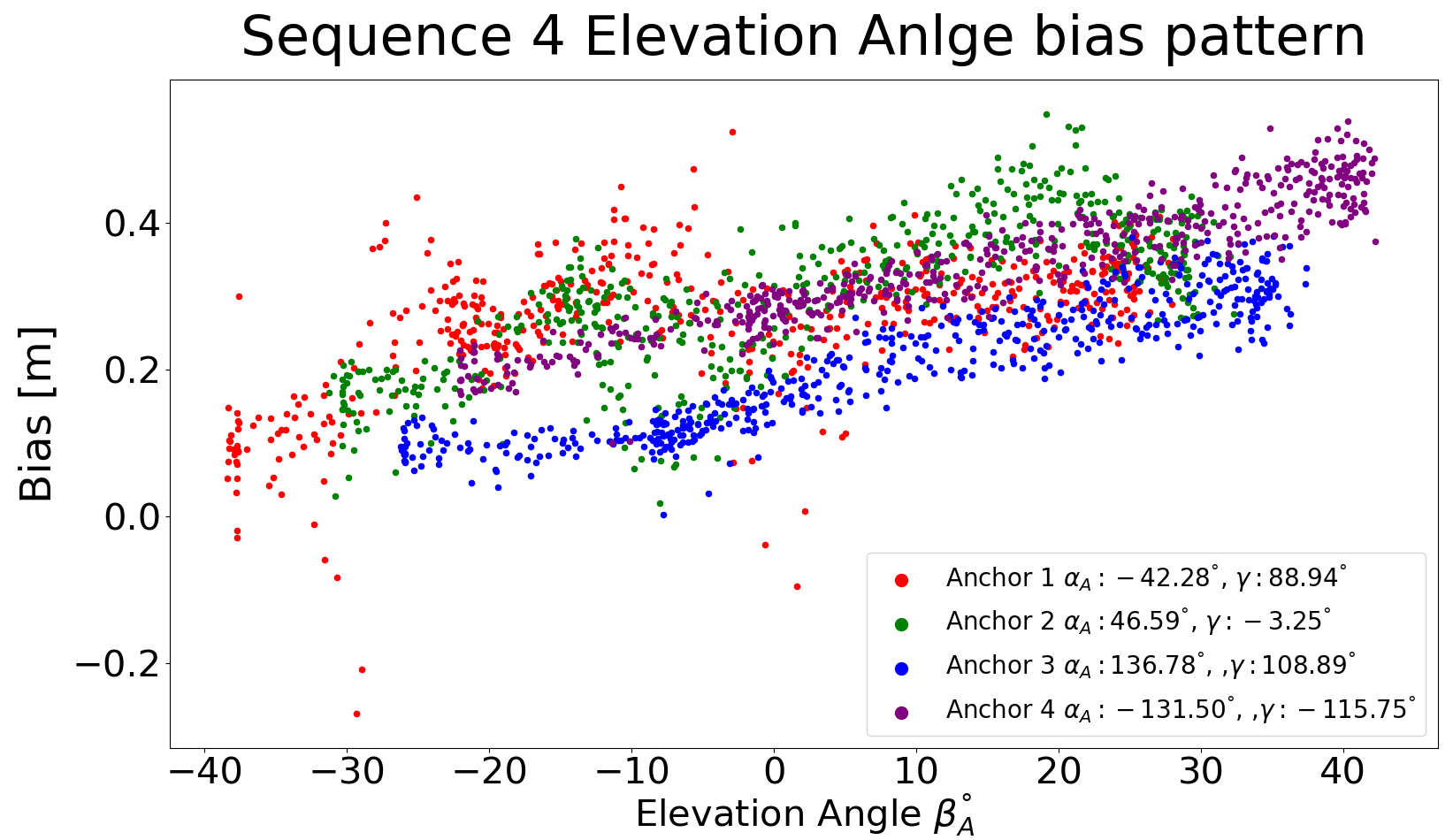}};
      \node[above left, inner sep=0em, xshift=2.5em, yshift=-0.5em] at (current bounding box.north west) {(d)};
    \end{tikzpicture} \\
    \begin{tikzpicture}
      \node[inner sep=0] at (0,0) {\includegraphics[width=0.45\textwidth, height=0.18\textheight]{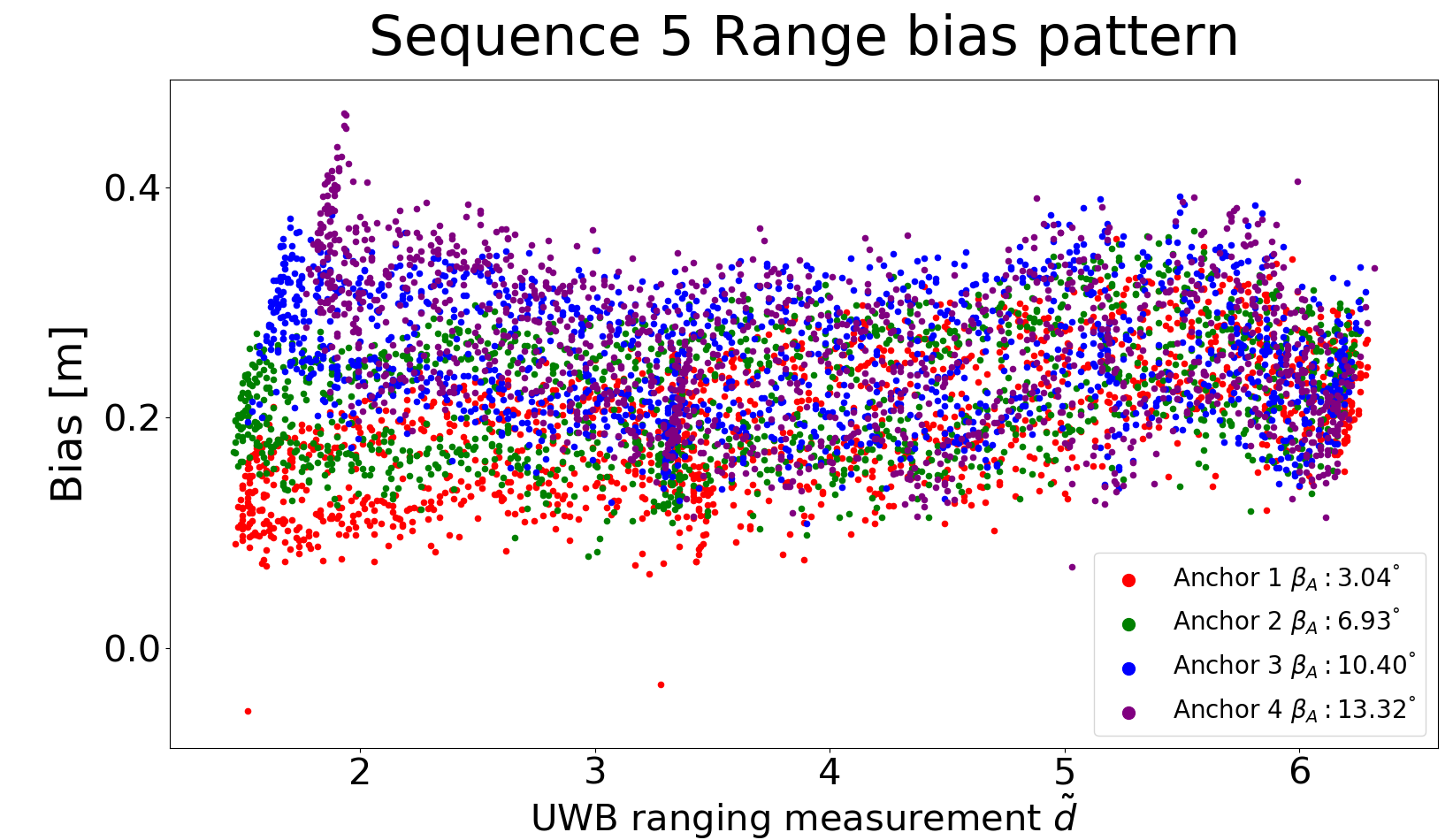}};
      \node[above left, inner sep=0em, xshift=2.5em, yshift=-0.5em] at (current bounding box.north west) {(e)};
    \end{tikzpicture} &
    \begin{tikzpicture}
      \node[inner sep=0] at (0,0) {\includegraphics[width=0.45\textwidth, height=0.18\textheight]{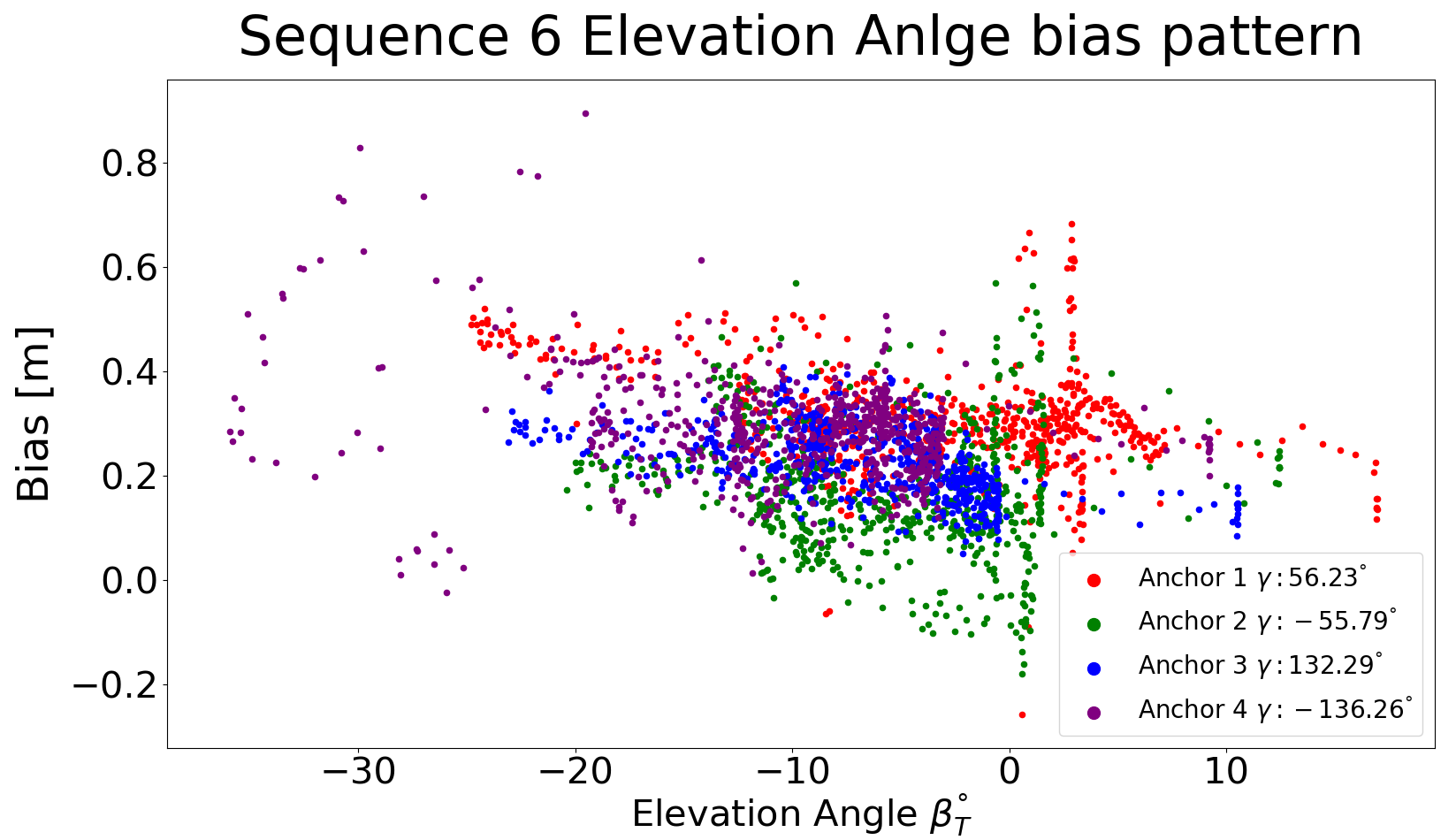}};
      \node[above left, inner sep=0em, xshift=2.5em, yshift=-0.5em] at (current bounding box.north west) {(f)};
    \end{tikzpicture} \\
  \end{tabular}
  \caption{ 
  UWB ranging bias patterns affected by bias factors ($\tilde{d}$ ,$\alpha$, $\beta$, $\gamma$) across different sequences. Each sequence focuses on one bias factor for analysis.
  }
  \label{analysis}
\end{figure*}

\begin{itemize}
    \item \textbf{Sequence 1} primarily addresses $\alpha_A$ and $\alpha_T$. The trajectory around the four anchors is designed to cover the full range of angles for $\alpha_A$ and $\alpha_T$.
    Additionally, to effectively decouple the impact of ranging measurement caused by the difference angle $\gamma$, we maintain a consistent heading direction for the drone. The trajectory includes two distinct altitudes while minimizing changes in the values of $\beta_A$ and $\beta_T$.
    
    \item \textbf{Sequence 2} aims to analyze the impact of a narrow range of $\alpha_A$, $\alpha_T$, and $\tilde{d}$. To achieve this, the four anchors are arranged in a line with the same heights. The drone's trajectory is set to move parallel to this anchor line to capture variations in $\alpha_A$,~$\alpha_B$. In the latter part of the sequence, the drone moves closer to the anchor line compared to the earlier part to examine the impact of a wider range in $\alpha$ and distance.
    
    \item \textbf{Sequence 3} involves the circular trajectory while continuously changing the drone's heading to address the impact of $\gamma$.
    Specifically, the drone's heading direction changes according to the tangent direction of the circular path.
    To reduce the impact of other factors except for $\tilde{d}$, we place the four anchors far from the trajectory, which reduced the range of $\alpha_A$,~$\alpha_B$,~$\beta_A$,~$\beta_B$.
    
    \item \textbf{Sequence 4} focuses on $\beta_A$ and $\beta_T$ by varying only the drone's altitude from the origin to completely decouple the impacts of other factors ($\alpha_A$,~$\alpha_B$,~$\gamma$). The four anchors are placed at different heights, expanding the elevation angle range $\beta$.
    
    \item \textbf{Sequence 5} is designed to address impact of $\tilde{d}$ while isolating other impact $\gamma$. We position the four anchors in a line, similar to Sequence 2, and place them very close to each other at different heights. The trajectory is designed to repeatedly vary the distance between the tag and the four anchors, making the distance alternately decrease and increase. Consequently, $\alpha$ and $\beta$ exhibit variations within a small range.
    
    \item \textbf{Sequence 6} is designed with a complex trajectory to capture the impact of all factors except $\gamma$, including $\alpha_A$,~$\alpha_A$,~$\beta_A$ and $\beta_B$. The drone moves in a complex manner within the internal area formed by the four anchors. 
\end{itemize}

\section{Data analysis}
We analyze the error and bias pattern of UWB range measurements according to the provided dataset sequences, focusing on decoupled impact factor $\alpha_A$, $\alpha_T$, $\beta_A$, $\beta_T$, $\gamma$ and $\tilde{d}$. We define a bias model function $b(\tilde{d},\bm{\alpha} ,\bm{\beta}, \bm{\gamma})$ in terms of UWB ranging measurement $\tilde{d}$, azimuth angle $\bm{\alpha} = [\alpha_A,\alpha_T]^T$, elevation angle $\bm{\beta} = [\beta_A,\beta_T]^T$ and difference angle $\bm{\gamma}$ similar to what is mentioned in ~\cite{LearningBase,Cal_MAP}. In the analysis, the UWB ranging measurement model can be described as follows:
\begin{align}
&\tilde{d} = \bm{d} + b(\tilde{d}, \bm{\alpha}, \bm{\beta}, \bm{\gamma}) + n, \\
&b(\tilde{d}, \bm{\alpha}, \bm{\beta}, \bm{\gamma}) = \tilde{d} - \bm{d}.
\label{ranging}
\end{align}
where $\tilde{d}$ is the measured distance by the UWB sensor, $\bm{d}$ is the true distance between the tag and the anchor, $b(\bm{d}, \bm{\alpha}, \bm{\beta}, \bm{\gamma})$ represents the bias pattern model and $n$ denotes the zero-mean Gaussian measurement noise. Bias is derived from Eq.~(\ref{ranging}).

The bias patterns resulting from the UWB relative pose are shown in Fig.~\ref{analysis}
We analyze the bias patterns related to $\alpha_A$ in Fig.~\ref{analysis}~(a). Due to the characteristics of the trajectory in Sequence 1, there are points where the tag's position closely aligns with the heading direction of the anchor. At this moment, $\alpha_A$ becomes close to $-180^\circ$ or $180^\circ$, resulting in a noticeable increase in the measurement bias. 

In contrast, Fig.~\ref{analysis}~(b) analyzes the bias factor $\alpha_T$. When the anchor is in the heading direction from the tag, when $\gamma$ is $-180^\circ$ or $180^\circ$, the bias is derived linearly. However, as $\alpha_T$ increases, a specific bias pattern becomes evident. 

In Fig.~\ref{analysis}~(c), the effect of the difference angle $\gamma$ on the bias pattern is observed. 
Regardless of the heading direction of the anchors, the circular trajectory allows for the measurement of $\gamma$ across the full range. A fairly consistent bias pattern can be identified by analyzing $\gamma$ for each anchor and tag. 

By fixing other bias factors~($\bm{\alpha}$, $\bm{\gamma}$) and varying $\beta_A$ within the range of $[-40^\circ, 40^\circ]$, we analyze the bias pattern in Fig.~\ref{analysis}~(d). As $\beta_A$ approaches $-40^\circ$, the tag is positioned above the anchor. Conversely, as $\beta_A$ approaches $-40^\circ$, the tag is positioned below the anchor. With an increase in $\beta_A$, the bias trend can be observed to increase linearly. 

In Fig.~\ref{analysis}~(e), we analyze how changes in range affect the bias, finding that the range factor does not impact the bias. However, due to the characteristics of the angles, as the distance between the anchor and the tag increases, $\bm{\alpha}$ and $\bm{\beta}$ inevitably converge. Thus, as the $\tilde{d}$ increases, the impact of $\bm{\alpha}$ and $\bm{\beta}$ diminishes, leading the bias to converge. 

Fig.~\ref{analysis}~(f) shows how the factor $\beta_T$ affects bias patterns. It is difficult to identify some apparent patterns in this figure. This is because the sequence is not designed to independently isolate the bias factors in the trajectory.

As a result of our analysis, we identify that the dominant bias factors are $\bm{\alpha}$, $\bm{\beta}$, and $\bm{\gamma}$. These dominant bias factors contribute to errors in UWB ranging measurements, and if not properly addressed, these errors can degrade the performance of UWB-based localization systems. Therefore, our dataset will be crucial for addressing the challenges associated with the relative pose of UWB.

\section{Conclusion}
Our dataset is designed for more precise UWB-based localization and provides an analysis of ranging measurements according to the relative pose of UWBs. To achieve this, it is captured using a qualified motion caption system, and unique experiments are conducted to minimize the interference between the elements constituting the relative poses, as well as complex element experiments. Based on these features, we expect this dataset to be a significant resource in the field of UWB-based localization estimation. Future work will involve creating an outdoor dataset that includes adverse weather conditions to contribute further to localization advancements in robotics applications.

\newpage
\section*{Acknowledgment}
This research was supported by the Challengeable Future Defense Technology Research and Development Program through the Agency For Defense Development~(ADD) funded by the Defense Acquisition Program Administration~(DAPA) in 2023~(No. 915052101).

\bibliographystyle{IEEEtran}
\bibliography{IEEEabrv,reference}

\end{document}